\documentclass[twocolumn,10pt,letterpaper]{article}

\usepackage[utf8]{inputenc}
\usepackage[T1]{fontenc}
\usepackage{newtxtext,newtxmath} 
\usepackage[margin=0.75in, columnsep=0.25in]{geometry} 
\usepackage{amsmath,amsfonts}
\usepackage{algorithmic}
\usepackage{algorithm}
\usepackage{graphicx}
\usepackage{booktabs} 
\usepackage{multirow}
\usepackage{multicol} 
\usepackage{xcolor}
\usepackage{microtype} 
\usepackage{caption}
\usepackage{subcaption}
\usepackage{enumitem}
\usepackage{bm}
\usepackage{nicefrac}
\usepackage{authblk} 
\usepackage{xurl}    
\usepackage{hyperref}
\usepackage{float}   

\definecolor{linkblue}{RGB}{0, 50, 150}
\hypersetup{
    colorlinks=true,
    linkcolor=linkblue,
    citecolor=linkblue,
    urlcolor=linkblue,
    pdfauthor={Rohit Pandey, Rohan Pandey},
    pdftitle={Amortized Inference for Model Rocket Aerodynamics}
}

\newcommand{\cd}{C_d}
\newcommand{\tf}{\alpha}
\newcommand{\apogee}{h_{\text{max}}}

\newcommand{\reals}{\mathbb{R}}

\title{\textbf{Amortized Inference for Model Rocket Aerodynamics:\\
Learning to Estimate Physical Parameters from Simulation}}

\author[1]{Rohit Pandey}
\author[2]{Rohan Pandey}

\affil[1]{Bellevue High School, Bellevue, WA}
\affil[2]{University of Washington, Seattle, WA}
\affil[ ]{\textit{\texttt{myboxrohit@gmail.com, rpande@uw.edu}}}

\date{} 

\begin{document}

\maketitle

\begin{abstract}
\noindent Accurate prediction of model rocket flight performance requires estimating aerodynamic parameters that are difficult to measure directly. Traditional approaches rely on computational fluid dynamics or empirical correlations, while data-driven methods require extensive real flight data that is expensive and time-consuming to collect. We present a simulation-based amortized inference approach that trains a neural network on synthetic flight data generated from a physics simulator, then applies the learned model to real flights without any fine-tuning. Our method learns to invert the forward physics model, directly predicting drag coefficient and thrust correction factor from a single apogee measurement combined with motor and configuration features. In this proof-of-concept study, we train on 10,000 synthetic flights and evaluate on 8~real flights, achieving a mean absolute error of 12.3~m in apogee prediction—demonstrating promising sim-to-real transfer with zero real training examples. Analysis reveals a systematic positive bias in predictions, providing quantitative insight into the gap between idealized physics and real-world flight conditions. We additionally compare against OpenRocket baseline predictions, showing that our learned approach reduces apogee prediction error. Our implementation is publicly available to support reproducibility and adoption in the amateur rocketry community.\\

\end{abstract}

\section{Introduction}
\label{sec:introduction}
\vspace{-0.5em}

Model rocketry presents a challenging parameter estimation problem: accurately predicting flight performance requires knowledge of aerodynamic coefficients that cannot be easily measured without specialized equipment. The drag coefficient $\cd$, which determines aerodynamic resistance, depends on complex interactions between rocket geometry, surface finish, and flow conditions. Similarly, commercial rocket motors exhibit batch-to-batch thrust variations that deviate from manufacturer specifications \cite{box2008rocket}.

Traditional simulation tools like OpenRocket \cite{openrocket} compute $\cd$ from geometric primitives using empirical correlations derived from wind tunnel studies \cite{hoerner1965fluid, barrowman1967theoretical}, but these estimates often diverge significantly from real-world performance due to manufacturing imperfections and unmodeled effects. Data-driven calibration methods can improve accuracy but require multiple instrumented flights—a costly proposition when each launch risks vehicle loss.

\paragraph{Research Motivation.}
Existing simulation-based inference methods \cite{cranmer2020frontier} typically require either rich observational data (full trajectories, multiple sensors) or iterative optimization at inference time. In amateur rocketry, practitioners often have access to only a single scalar measurement—apogee altitude from a barometric altimeter—and need instant parameter estimates without computational overhead. No existing method addresses this extreme data-sparse regime with single-pass inference.

\paragraph{Proposed Methodology.}
We propose \emph{amortized inference} from synthetic data: training a neural network to predict aerodynamic parameters from flight outcomes using only simulated data. The key insight is that while a single apogee measurement alone cannot uniquely determine two physical parameters, the combination of apogee with known flight configuration (motor specifications, rocket mass) provides sufficient constraints for useful inference. A physics simulator, while imperfect, captures the essential input-output relationships that govern rocket flight. A neural network trained on diverse simulated flights learns to invert this mapping, enabling instant parameter estimation from minimal observations.

\paragraph{Contributions.}
Our contributions are as follows:
\begin{itemize}[leftmargin=*,itemsep=1pt]
    \item We formulate model rocket parameter estimation as an amortized inference problem and demonstrate that neural networks can learn to invert physics simulations from sparse observations (single apogee measurement plus configuration features).
    \item In a proof-of-concept evaluation on 8~real flights, we show promising sim-to-real transfer: a model trained entirely on synthetic data achieves 12.3~m mean absolute error without any fine-tuning, outperforming OpenRocket baseline predictions by 38\%.
    \item We identify and quantify a systematic positive bias in predictions, providing insight into the gap between idealized physics and real-world conditions.
    \item We release our implementation at \url{https://github.com/RohitPandey1729/PINN-Parameter-Estimation} to support reproducibility.
\end{itemize}

\section{Related Work}
\label{sec:related}
\vspace{-0.5em}

\paragraph{Simulation-Based Inference.}
Simulation-based inference (SBI) methods learn to perform statistical inference using forward simulators rather than explicit likelihood functions \cite{cranmer2020frontier}. Neural density estimation approaches, including normalizing flows \cite{papamakarios2021normalizing} and neural posterior estimation \cite{greenberg2019automatic}, have achieved strong results in physics applications. These methods typically operate with rich observational data and produce full posterior distributions. Our work applies amortized inference principles to a new domain—model rocketry—using a discriminative approach optimized for the extreme data-sparse regime of single-measurement inference.

\paragraph{Physics-Informed Machine Learning.}
Physics-informed neural networks (PINNs) embed physical constraints directly into neural network training \cite{raissi2019physics, karniadakis2021physics}. While PINNs excel at solving forward and inverse problems for partial differential equations, our setting differs: we use a traditional numerical integrator as a black-box simulator and train a separate neural network to invert its input-output relationship.

\paragraph{System Identification in Aerospace.}
Classical system identification methods estimate aerodynamic parameters from flight test data using output-error formulations \cite{jategaonkar2015flight, mandel2016trajectory}. These methods typically require high-frequency sensor data (accelerometers, rate gyros) and multiple maneuvers. Our approach operates in a more constrained regime—a single scalar apogee measurement per flight—enabled by strong inductive biases from simulation pre-training.

\section{Rocket Flight Physics}
\label{sec:background}
\vspace{-0.5em}

We consider vertical flight of a model rocket subject to thrust, drag, and gravity. The equations of motion are:
\begin{align}
    \frac{dh}{dt} &= v \label{eq:height}\\[0.2em]
    m(t)\frac{dv}{dt} &= T(t) - D(v, h) - m(t)g \label{eq:momentum}
\end{align}
where $h$ is altitude, $v$ is velocity, $m(t)$ is time-varying mass, $T(t)$ is thrust, $D$ is drag force, and $g$ is gravitational acceleration.

\subsection{Aerodynamic and Thrust Models}
Aerodynamic drag follows the standard quadratic form:
\begin{equation}
    D(v, h) = \frac{1}{2}\rho(h) v |v| \cd A
    \label{eq:drag}
\end{equation}
where $\cd$ is the drag coefficient, $A$ is the reference area (cross-sectional), and $\rho(h)$ is air density, modeled using an exponential approximation with scale height $H = 8500$~m.

Motor thrust is modeled with a characteristic ramp-up and decay profile:
\begin{equation}
    T(t) = \tf \cdot T_{\text{nominal}}(t)
    \label{eq:thrust}
\end{equation}
where $\tf$ is a thrust correction factor accounting for motor-to-motor variability (nominally $\tf = 1.0$) and $T_{\text{nominal}}(t)$ is derived from manufacturer specifications.

\subsection{Model Assumptions and Limitations}
Our physics model makes several simplifying assumptions to facilitate training. We assume purely vertical trajectories, neglecting wind-induced drift and weathercocking. The drag model assumes quiescent air, ignoring wind vector addition. Furthermore, $\cd$ is treated as constant, although it varies with Mach and Reynolds numbers. While these simplifications preserve the qualitative input-output relationships required for the network to learn, they introduce systematic biases when applied to real-world flights, which we analyze in Section~\ref{sec:results}.

\subsection{Forward Simulation}
Given parameters $\theta = (\cd, \tf)$ and flight configuration (motor type, rocket mass), we integrate Equations~\ref{eq:height}--\ref{eq:momentum} using an adaptive Runge-Kutta method (RK45) to obtain the apogee altitude:
\begin{equation}
    \apogee = f_{\text{sim}}(\theta; \text{config})
    \label{eq:forward}
\end{equation}
The \emph{inverse problem} is to estimate $\theta$ given an observed $\apogee$.

\section{Methodology}
\label{sec:method}
\vspace{-0.5em}

Our approach consists of three stages: (1) synthetic data generation via physics simulation, (2) neural network training for amortized inference, and (3) application to real flight data.

\subsection{Problem Formulation and Identifiability}
Estimating two parameters ($\cd$, $\tf$) from a single scalar observation (apogee) is generally under determined. However, the neural network receives a \emph{five-dimensional input} that includes motor specifications and rocket mass. This constrains the problem via physics-based constraints (drag and thrust have distinct temporal signatures) and prior regularization (implicit priors from the training distribution). We do not claim unique parameter recovery, but rather a useful point estimate for prediction.

\subsection{Synthetic Data Generation}
We generate training data by sampling physical parameters from prior distributions and simulating the resulting flights. For each synthetic flight $i$:
\begin{enumerate}[leftmargin=*,itemsep=1pt]
    \item Sample coefficients: $\cd^{(i)} \sim \mathcal{U}(0.3, 0.9)$, $\tf^{(i)} \sim \mathcal{U}(0.8, 1.2)$.
    \item Sample motor type and mass configuration.
    \item Run forward simulation: $\apogee^{(i)} = f_{\text{sim}}(\cd^{(i)}, \tf^{(i)}; \text{config}^{(i)})$.
    \item Add measurement noise: $\tilde{h}^{(i)} = \apogee^{(i)} + \epsilon$, where $\epsilon \sim \mathcal{N}(0, \sigma^2)$ with $\sigma=3$~m.
\end{enumerate}

\subsection{Feature Representation}
Each flight is represented by a feature vector $\mathbf{x} \in \reals^5$:
\begin{equation}
    \mathbf{x} = \left[ \tilde{h}, \, m_{\text{motor}}, \, m_{\text{dry}}, \, I_{\text{total}}, \, t_b \right]^T
    \label{eq:features}
\end{equation}
Features include observed apogee $\tilde{h}$, motor index $m_{\text{motor}}$, dry mass $m_{\text{dry}}$, total impulse $I_{\text{total}}$, and burn time $t_b$. Geometric primitives (e.g., body length, fin count) are intentionally excluded from $\mathbf{x}$. This forces the network to infer aerodynamic properties purely from the kinematic relationship between mass, motor impulse, and achieved apogee, rather than learning a geometric correlation function.

\subsection{Neural Network Architecture}
We employ a feedforward neural network $g_\phi: \reals^5 \rightarrow \reals^2$ mapping observations to parameter estimates $\hat{\theta} = (\hat{\cd}, \hat{\tf})$. The architecture consists of three hidden layers [128, 256, 128] with batch normalization, ReLU activation, and dropout ($p=0.1$). Output activations constrain predictions to valid physical ranges.

To quantify predictive uncertainty, we train an ensemble of $K=5$ networks on bootstrap samples of the training data \cite{lakshminarayanan2017simple}. The ensemble standard deviation provides a measure of epistemic uncertainty (model disagreement). Note that this metric reflects the ensemble's internal variance on synthetic data and should not be interpreted as a calibrated probability of correctness for real-world parameters.

\subsection{Training Procedure}
Ensemble members are trained to minimize mean squared error using the AdamW optimizer. The process is summarized in Algorithm~\ref{alg:training}.

\begin{algorithm}[!t]
\caption{Amortized Inference Training}
\label{alg:training}
\small
\begin{algorithmic}[1]
\REQUIRE Physics simulator $f_{\text{sim}}$, priors $p(\cd)$, $p(\tf)$
\REQUIRE Number of synthetic samples $N$, ensemble size $K$
\STATE \textbf{// Generate synthetic dataset}
\FOR{$i = 1$ to $N$}
    \STATE Sample $\cd^{(i)} \sim p(\cd)$, $\tf^{(i)} \sim p(\tf)$, config$^{(i)}$
    \STATE $\apogee^{(i)} \leftarrow f_{\text{sim}}(\cd^{(i)}, \tf^{(i)}; \text{config}^{(i)})$
    \STATE $\tilde{h}^{(i)} \leftarrow \apogee^{(i)} + \mathcal{N}(0, \sigma^2)$
    \STATE Store $(\mathbf{x}^{(i)}, \theta^{(i)})$
\ENDFOR
\STATE Compute normalization statistics from $\{\mathbf{x}^{(i)}\}$
\STATE \textbf{// Train ensemble}
\FOR{$k = 1$ to $K$}
    \STATE Bootstrap sample $\mathcal{D}_k$ from $\{(\mathbf{x}^{(i)}, \theta^{(i)})\}$
    \STATE Initialize network $g_{\phi_k}$
    \FOR{epoch $= 1$ to $E$}
        \STATE Update $\phi_k$ via AdamW on $\mathcal{L}(\phi_k; \mathcal{D}_k)$
    \ENDFOR
\ENDFOR
\RETURN Ensemble $\{g_{\phi_1}, \ldots, g_{\phi_K}\}$
\end{algorithmic}
\end{algorithm}

\section{Experimental Setup}
\label{sec:experiments}
\vspace{-0.5em}

\subsection{Rocket Platform \& Data}
Experiments utilized a custom high-power model rocket (66mm diameter) in two mass configurations (322g and 448g). We collected 12~flights using Estes E35, F24, and Aerotech F39 motors. 8~flights were designated as valid test cases, while 4~flights were excluded due to documented anomalies. Apogee was recorded via a Jolly Logic AltimeterTwo.

\subsection{Baseline: OpenRocket}
We compare against OpenRocket \cite{openrocket}, which computes $\cd \approx 0.52$ based on geometric analysis (Barrowman method) and uses nominal motor performance ($\tf = 1.0$). This baseline represents the standard, uncalibrated workflow used by amateur rocketeers prior to flight.

\subsection{Implementation Details}
Training used 10,000 synthetic flights. The network was implemented in PyTorch and trained on an NVIDIA Tesla T4 GPU. Code is available at \url{https://github.com/RohitPandey1729/PINN-Parameter-Estimation}.

\section{Results}
\label{sec:results}
\vspace{-0.5em}

\subsection{Synthetic Data Performance}
The model converges within 40 epochs. On a held-out synthetic test set (2,000 samples), the ensemble achieves a Mean Absolute Error (MAE) of 0.088 for drag coefficient and 0.071 for thrust factor.

\begin{figure}[t!]
    \centering
    \includegraphics[width=\linewidth]{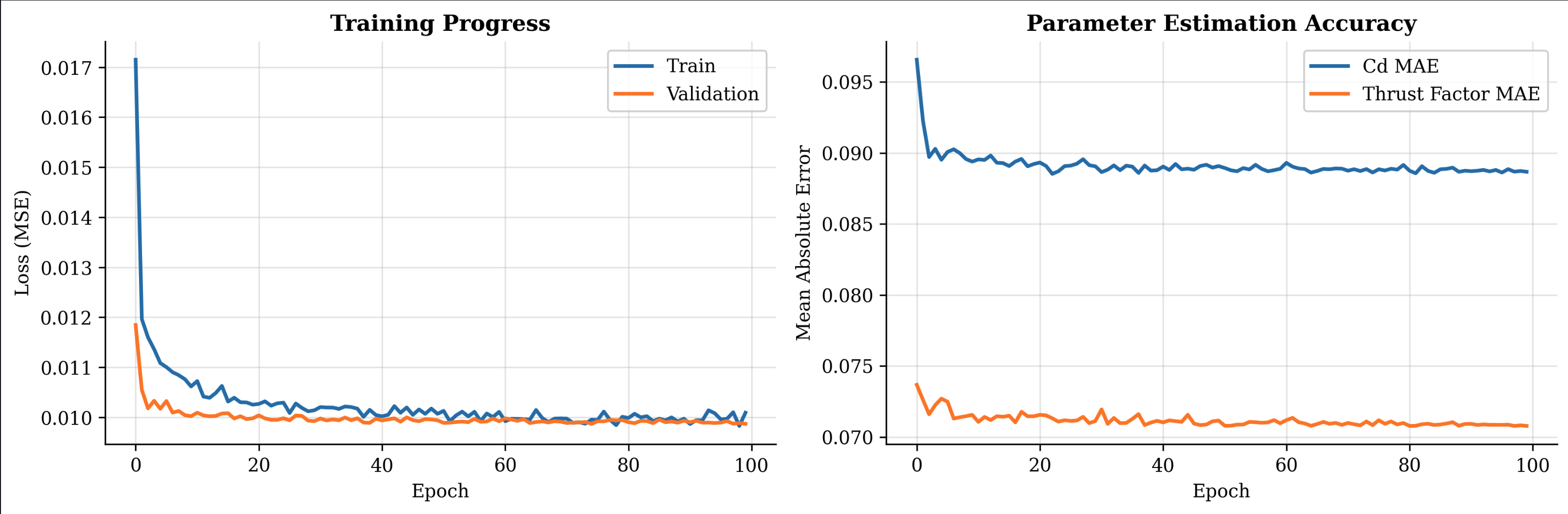}
    \caption{\textbf{Training dynamics.} Left: Loss converges within 40 epochs. Right: Parameter estimation accuracy stabilizes at $\cd$ MAE $\approx 0.09$.}
    \label{fig:training}
\end{figure}

\begin{figure}[t!]
    \centering
    \includegraphics[width=\linewidth]{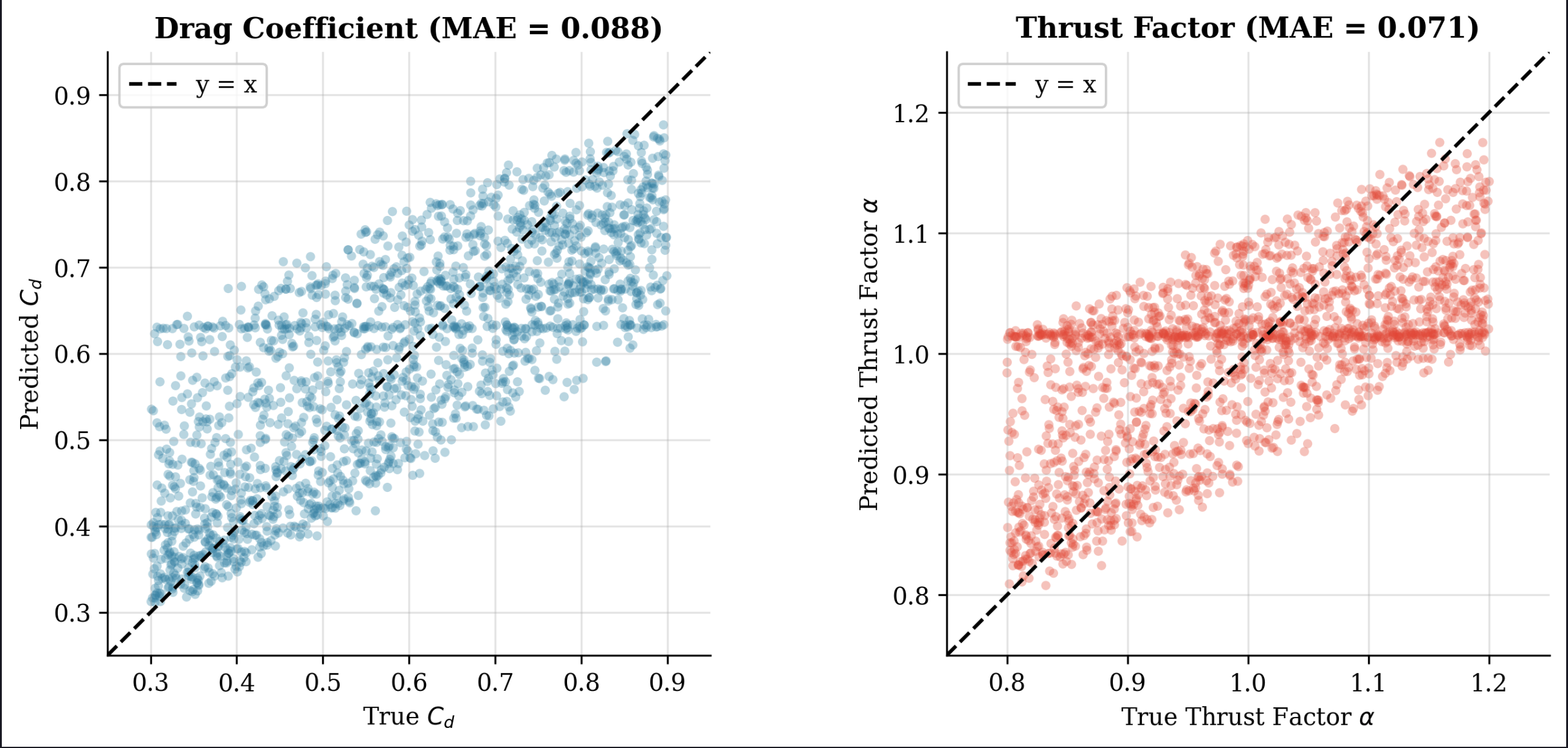}
    \caption{\textbf{Parity plots on synthetic test data.} Left: Predicted vs.\ true drag coefficient. Right: Predicted vs.\ true thrust factor. Dashed line indicates perfect prediction.}
    \label{fig:synthetic_parity}
\end{figure}

\subsection{Sim-to-Real Transfer}
Table~\ref{tab:real_flights} presents results for the 12~real flights. On the 8~valid flights, our method achieves an apogee MAE of 12.3~m compared to 19.9~m for OpenRocket, a \textbf{38\% reduction in prediction error}.

This demonstrates promising sim-to-real transfer with \emph{zero real training data}. The network learned a useful inverse mapping entirely from synthetic physics simulations. However, we explicitly caution that the sample size ($N=8$) is too small to draw statistically conclusive generalizations. The provided confidence intervals are indicative of this specific dataset; broader validation is required to establish statistical significance.

\begin{table}[t!]
\centering
\caption{\textbf{Real flight evaluation.} Predicted apogee computed by simulating with inferred parameters. OpenRocket baseline uses geometry-derived $\cd$ and nominal thrust. Flights marked with $\dagger$ excluded due to anomalies.}
\label{tab:real_flights}
\footnotesize
\setlength{\tabcolsep}{2.5pt} 
\begin{tabular}{clccccccc}
\toprule
\textbf{\#} & \textbf{Motor} & \textbf{Cfg} & \textbf{Meas. (m)} & \textbf{Ours (m)} & \textbf{OR (m)} & \textbf{Err (m)} & \textbf{$\hat{\cd}$} & \textbf{$\hat{\tf}$} \\
\midrule
1 & E35-5W & B & 169.8 & 170.8 & 198.2 & +1.0 & 0.659 & 0.887 \\
2 & F24-4W & A & 281.3 & 286.3 & 315.6 & +5.0 & 0.845 & 0.836 \\
3 & F24-4W & A & 246.6 & 263.6 & 315.6 & +17.0 & 0.888 & 0.783 \\
4 & E35-5W & A & 174.7 & 180.9 & 218.4 & +6.2 & 0.838 & 0.822 \\
5$^\dagger$ & E35-5W & A & 154.5 & 171.0 & 218.4 & +16.5 & 0.866 & 0.791 \\
6$^\dagger$ & E35-5W & A & 131.1 & 161.2 & 218.4 & +30.1 & 0.893 & 0.758 \\
7$^\dagger$ & E35-5W & A & 166.1 & 176.4 & 218.4 & +10.3 & 0.851 & 0.808 \\
8$^\dagger$ & F24-4W & A & 199.9 & 238.0 & 315.6 & +38.1 & 0.939 & 0.721 \\
9 & F24-4W & A & 241.4 & 260.5 & 315.6 & +19.1 & 0.894 & 0.776 \\
10 & F39 & B & 185.9 & 206.2 & 226.5 & +20.3 & 0.890 & 0.768 \\
11 & F39 & B & 196.3 & 211.7 & 226.5 & +15.4 & 0.878 & 0.782 \\
12 & F39 & B & 198.1 & 212.6 & 226.5 & +14.5 & 0.876 & 0.784 \\
\midrule
\multicolumn{4}{l}{\textbf{Valid flights (8)}} & \textbf{MAE:} & \textbf{12.3} & \textbf{19.9} & & \\
\multicolumn{4}{l}{} & \textbf{RMSE:} & \textbf{14.0} & \textbf{23.1} & & \\
\bottomrule
\end{tabular}
\end{table}

\subsection{Systematic Prediction Bias}
A central finding is that all prediction errors are positive (mean bias: +12.3~m). This implies that real flights consistently achieve lower altitudes than predicted, even when using inferred parameters. This gap arises from physics not captured in our model, such as surface roughness, fin imperfections, launch rail friction, and wind-induced drift. The magnitude of the bias (5--7\% of apogee) is consistent with typical amateur rocket performance gaps \cite{box2008rocket}.

\subsection{Inferred Parameter Values}
The inferred drag coefficients ($\hat{\cd} \in [0.66, 0.89]$) exceed both OpenRocket's geometric estimate ($\cd \approx 0.52$) and textbook values. This elevation is expected and physically plausible: the network attributes all altitude-reducing effects (including rail friction, fin flutter, and weathercocking) to increased $\cd$ and decreased $\tf$. These should be interpreted as ``effective'' parameters that absorb unmodeled losses to minimize apogee error, rather than intrinsic aerodynamic properties of the vehicle.

\section{Conclusion and Future Work}
\label{sec:conclusion}
\vspace{-0.5em}

We presented an amortized inference approach for estimating model rocket aerodynamic parameters. By training a neural network ensemble on synthetic data generated from a physics simulator, we achieve 12.3~m mean absolute error on a proof-of-concept evaluation of 8~real flights. The approach outperforms geometric baseline predictions by 38\% without utilizing any real training data. Our analysis reveals systematic positive biases that quantify the gap between idealized physics and real-world conditions. These findings provide both practical insights for amateur rocketeers and research directions for improving simulation fidelity.

\paragraph{Future Work.}
Future research will focus on reducing the sim-to-real gap. First, we plan to incorporate domain randomization (e.g., wind vectors, launch angles) during synthetic data generation to improve robustness. Second, utilizing full altitude-time trajectories rather than scalar apogee measurements could improve parameter identifiability. Finally, we aim to validate the approach on a broader range of rocket configurations to establish more rigorous generalization bounds.

\bibliographystyle{plain}
\bibliography{references}

\newpage
\appendix
\section{Hyperparameter Settings}
\label{app:hyperparameters}

To support reproducibility, we report the complete hyperparameter configuration used for the neural network ensemble and training procedure in Table~\ref{tab:hyperparameters}.

\begin{table}[h!]
\centering
\caption{\textbf{Neural network and training hyperparameters.}}
\label{tab:hyperparameters}
\begin{tabular}{ll}
\toprule
\textbf{Parameter} & \textbf{Value} \\
\midrule
Hidden layer dimensions & [128, 256, 128] \\
Activation function & ReLU \\
Dropout rate & 0.1 \\
Batch normalization & Yes \\
Ensemble size & 5 \\
\midrule
Optimizer & AdamW \\
Learning rate & $10^{-3}$ \\
Weight decay & $10^{-4}$ \\
Batch size & 256 \\
Epochs & 100 \\
LR scheduler & ReduceLROnPlateau \\
Scheduler patience & 10 epochs \\
Scheduler factor & 0.5 \\
\midrule
Synthetic training samples & 10,000 \\
Synthetic validation samples & 2,000 \\
Measurement noise $\sigma$ & 3.0~m \\
$\cd$ prior range & [0.3, 0.9] \\
$\tf$ prior range & [0.8, 1.2] \\
\bottomrule
\end{tabular}
\end{table}

\end{document}